\documentclass{article}
\usepackage[nonatbib]{neurips_2020}
\usepackage[utf8]{inputenc} % allow utf-8 input
\usepackage[T1]{fontenc}    % use 8-bit T1 fonts
\usepackage{hyperref}       % hyperlinks
\usepackage{url}            % simple URL typesetting
\usepackage{booktabs}       % professional-quality tables
\usepackage{amsfonts}       % blackboard math symbols
\usepackage{nicefrac}       % compact symbols for 1/2, etc.
\usepackage{microtype}      % microtypography
\usepackage{mathtools}
\newcommand{\pril}{\textsc{PRIL}} 
\newtheorem{definition}{Definition}
\usepackage{amsmath}
\usepackage{multirow}
\usepackage{xcolor}

\title{How Private Is Your RL Policy? An Inverse RL Based Analysis Framework}
\date{December 2021}
\author{ 
    Kritika Prakash \\ Machine Learning Lab \\ IIIT Hyderabad \\
	\texttt{kritika.prakash@research.iiit.ac.in} \\
	\And
    Fiza Husain \\ Machine Learning Lab \\ IIIT Hyderabad \\
	\texttt{fiza.husain@students.iiit.ac.in} \\
	\AND
    Praveen Paruchuri \\ Machine Learning Lab \\ IIIT Hyderabad \\
	\texttt{praveen.p@iiit.ac.in} \\
	\And
    Sujit P. Gujar \\ Machine Learning Lab \\ IIIT Hyderabad \\
	\texttt{sujit.gujar@iiit.ac.in} \\
}
\hypersetup{
pdftitle={How Private Is Your RL Policy?},
pdfsubject={},
pdfauthor={Kritika Prakash, Fiza Husain, Praveen Paruchuri, Sujit P. Gujar},
pdfkeywords={differential privacy, reinforcement learning, inverse RL, adversarial attack},
}
\begin{document}
\maketitle
\begin{abstract}
{Reinforcement Learning (RL) enables agents to learn how to perform various tasks from scratch. In domains like autonomous driving, recommendation systems and more, optimal RL policies learned could cause a privacy breach if the policies memorize any part of the private reward. We study the set of existing differentially-private RL policies derived from various RL algorithms such as Value Iteration, Deep Q Networks, and Vanilla Proximal Policy Optimization. We propose a new Privacy-Aware Inverse RL (\pril) analysis framework, that performs reward reconstruction as an adversarial attack on private policies that the agents may deploy. For this, we introduce the reward reconstruction attack, wherein we seek to reconstruct the original reward from a privacy-preserving policy using an Inverse RL algorithm. An adversary must do poorly at reconstructing the original reward function if the agent uses a tightly private policy. Using this framework, we empirically test the effectiveness of the privacy guarantee offered by the private algorithms on multiple instances of the FrozenLake domain of varying complexities. Based on the analysis performed, we infer a gap between the current standard of privacy offered and the standard of privacy needed to protect reward functions in RL. We do so by quantifying the extent to which each private policy protects the reward function by measuring distances between the original and reconstructed rewards.}
\end{abstract}
% \keywords{Differential Privacy \and Reinforcement Learning \and Inverse RL \and Adversarial Attack}

\section{Introduction}
%%%%%%%%%%%%%%%%%%%%%%%%% Figure 1 %%%%%%%%%%%%%%%%%%%%%%%%%%%
\begin{figure*}[h]
\centering
\includegraphics[scale=0.45]{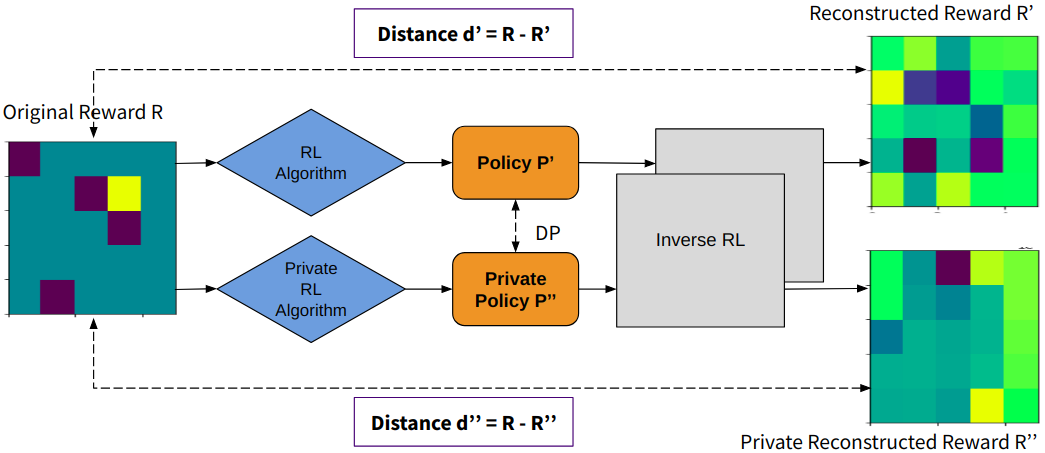}
\caption{\pril: Privacy-Aware Inverse RL analysis framework}
\label{fig:1_flowchart}
\end{figure*}
%%%%%%%%%%%%%%%%%%%%%%%%%%%%%%%%%%%%%%%%%%%%%%%%%%%%%%%%%%%%%%

Recent advancements in reinforcement learning (RL) have found widespread application in many real-world domains. Often, these domains are built from rich data sources or real-world environments, which could contain sensitive information of many individuals. This is evident in domains such as autonomous driving, recommendation systems, trading, industrial assembly, and domestic service robots. For example, a recommendation system agent for an online shopping platform not only tracks the purchases made, but also how long the user hovers over an item that he/she did not purchase. Another example is when an autonomous driving agent not only learns the dynamics behind driving, but also identifies people and predicts their behaviours on the streets, and how to respond to such situations. The reward function in these environments is often sensitive, as it is built on people's private information. RL agents trained in such environments should not expose the private information of individuals. We therefore, need to use privacy-preserving methods to protect the rewards from being memorized in the agent's policy in such a manner that the agent's utility is not compromised. 

Recent works use \emph{Differential Privacy} (DP) \cite{10.1007/11681878_14} to make the agent's policy quantifiably private and to get a rigorous privacy guarantee. Like many other areas in AI, RL has also started adopting DP to establish a mathematical way of guaranteeing data privacy in RL environments. However, an important question is: does the privacy guarantee offered by the private policies translate well into protecting the reward function? If not, how can we understand the gap in achieving meaningful privacy? What role does the type of reward function, algorithms (both RL and DP algorithms), and environment play? In other words, does privacy in policy translate well to privacy in reward? To investigate this, we first build an autonomous agent whose aim is to learn a private policy using existing privacy techniques. The intent is to reach a goal state that helps maximize the agent's expected reward in discrete finite-state environments. We then investigate the true level of reward privacy offered by the existing state-of-the-art privacy techniques for RL algorithms.

We evaluate the private policies by estimating an adversary's ability to reconstruct the original reward function from the agent's learned policy. The field of Inverse RL \cite{ng2000algorithms} arose to solve this exact problem of extracting reward signal given the observed, optimal behaviour. Given the ``reverse engineering" nature of inverse RL, the reconstructed reward can be used as an adversarial attack on environments with protected rewards. Building on this key intuition, we propose the \pril \ analysis framework that first performs the reward reconstruction attack, and then computes its similarity to the original private reward via various distance metrics to test the strength of the attack.

We apply this framework over a set of privacy preserving techniques: 
\begin{enumerate}
    \item Bellman update DP
    \item R\'enyi-DP in deep learning (DL)
    \item Functional noise DP
\end{enumerate}
These are applied to a set of RL algorithms:
\begin{enumerate}
    \item Deep Q Network (DQN)
    \item Vanilla Proximal Policy Optimization (PPO)
    \item Value Iteration (VI)
\end{enumerate} 
We build privacy into the agent from multiple perspectives: a DL perspective, an RL perspective, and a deep RL perspective. Through experiments, we show that there exists a gap between the privacy offered via the current private RL methods, and the privacy needed to protect reward functions. We also present the privacy-utility trade-off achieved by each policy. We show that privacy in policy does not translate to privacy in reward, as the reconstruction error is independent of the $\epsilon$-DP budget. DP based policies are unsuccessful at protecting the sensitive reward function due to a gap in privacy. Our experiments demonstrate that there is a need to further inspect the effectiveness of DP policies to protect sensitive reward functions. It is a serious privacy threat if an adversary is able to infer the rewards in spite of using a private policy.

%The key private element of any RL environment is its reward function. It helps understand the inherent value of being in a state (and taking an action). Sometimes, we have access to optimal policies for environments that do not have a well-defined reward structure. 
% does not get instantaneous feedback when reward is sparse 
% Sensitivity analysis of Bellman update in VI and of PPO.
% Integrating inverse RL with a rich deep RL environment  
% Adding privacy to existing RL environments and algorithms

In summary, our key contributions are:
\begin{enumerate}
    \item  We study and analyze the existing set of privacy techniques for RL.
    \item We introduce a novel reward reconstruction attack and supporting \pril \ framework.
    \item We empirically evaluate the performance of various private deep RL policies within our framework. 
    \item We identify and quantify the gap between the privacy offered in policy and the privacy needed in reward.
\end{enumerate}

The next section reviews related work. Section \ref{sec:3_prelim} provides background on RL, DP, and inverse RL. Section \ref{sec:4_inv_rl} explains the \pril \ framework. Section \ref{sec:5_exp_set_up} describes the experimental pipeline and setup. Section \ref{sec:6_discuss} provides an empirical analysis and discussion on our findings. Section \ref{sec:7_conclusion} concludes the work. Deferred proofs and scope appears in the appendix. All source code and experiments are made publicly available\footnote{Link to code: \url{https://github.com/magnetar-iiith/PRIL}}.

\section{Related Work}

Previous works on privacy-preserving RL make the use of DP. \cite{vietri2020private} shows how to achieve joint-DP in episodic-RL via the upper confidence bound and Q-learning algorithms, where each training episode comes from a different environment. \cite{wang2019privacy} makes the use of functional noise to make Q-learning private. \cite{vietri2020private} gives us probably-approximately correct (PAC) and regret guarantees for private RL. \cite{balle2016differentially} propose differentially private policy evaluation using the Monte-Carlo algorithm. \cite{hannun2019privacy} introduces a private way of using multi-party contextual bandits. For the actor-critic class of algorithms, \cite{lebensold2019actor} presents a differentially private critic, and \cite{seo2020differentially} presents a differentially private actor.
% TODO: Contrast this section against this work and point out other gaps in existing literature

In RL, a significant amount of work has been done to perform adversarial attacks that target the quality of the learned policy (\cite{gleave2019adversarial}, \cite{huang2017adversarial}, \cite{kos2017delving}, \cite{chen2019adversarial}). Building on this, there has been work to make RL policies robust to such attacks \cite{oikarinen2020robust}. However, very few works have looked into privacy attacks in RL. \cite{pan2019you} shows that agents can memorize the environment and its private transition dynamics, by performing privacy attacks using genetic algorithms and candidate inference. \cite{fu2017learning} introduce an adversarial inverse RL algorithm based on an adversarial reward learning formulation to improve robustness.

While these previous works tackle building privacy in policies using Differential Privacy, they do not investigate its impact on the underlying private data i.e., the reward function used to learn the policies. Many adversarial attacks such as the membership inference attack, linkage attack, and data-reconstruction attack have been used to evaluate the level of privacy attainable by a data-analysis system. We introduce a new privacy attack that targets the private reward function. Our proposed attack - the reward-reconstruction attack is a special case of the data-reconstruction attack. We use inverse RL to learn the reward, and to assess the quality of private RL algorithms.

\section{Background and Preliminaries}
\label{sec:3_prelim}
In this section, we introduce the basics of RL, inverse RL, and DP.
% \if 0
\begin{table}[t]
\centering
\renewcommand{\arraystretch}{1.3} 
\begin{tabular}{|p{2.4cm}|p{4.6cm}|}
\hline
 Abbreviation & Full Form \\
 \hline
 DP & Differential Privacy \\
 RL & Reinforcement Learning \\
 MDP & Markov Decision Process \\
 DL & Deep Learning \\
 LP & Linear Programming \\
 RDP & R\`enyi Differential Privacy \\
 PAC & Probably Approximately Correct \\
 \hline
 PRIL & Privacy-Aware Inverse RL\\
 VI & Value Iteration\\
 DQN & Deep Q Network\\
 PPO & Proximal Policy Optimization\\
 \hline
 DP-Bellman & Private Bellman update\\
 DP-SGD & Private SGD optimizer + ReLU\\
 DP-Shoe & Private SGD optimizer + tan-h\\
 DP-Adam & Private Adam optimizer + ReLU\\
 DP-FN & Private Functional Noise engine\\
 \hline
 VI-DP-Bellman & VI + DP-Bellman\\
 DQN-DP-SGD & DQN + DP-SGD\\
 DQN-DP-Shoe & DQN + DP-Shoe\\
 DQN-DP-Adam & DQN + DP-Adam\\
 DQN-DP-FN & DQN + DP-FN\\
 PPO-DP-SGD & PPO + DP-SGD actor\\
 PPO-DP-Shoe & PPO + DP-Shoe actor\\
 PPO-DP-Adam & PPO + DP-Adam actor\\
 \hline
\end{tabular}
\label{tab:acronyms}
\vspace{0.3cm}
\caption{List of acronyms used}
\end{table}
% \fi

\begin{table*}[t]
\centering
\renewcommand{\arraystretch}{1.3} 
\begin{tabular}{|p{2cm}|p{4cm}|p{4cm}|p{4cm}|}
\hline
 No. & Policy Class & RL Algorithm & Privacy Technique \\
 \hline
 1 & VI-DP-Bellman & Value Iteration & Bellman Update DP \\
 2 & DQN-DP-SGD & Deep Q Network & DP-SGD \\
 3 & DQN-DP-Shoe & Deep Q Network & DP-Shoe\\
 4 & DQN-DP-Adam & Deep Q Network & DP-Adam\\
 5 & DQN-DP-FN & Deep Q Network & DP-FN\\
 6 & PPO-DP-SGD & Vanilla PPO & DP-SGD actor\\
 7 & PPO-DP-Shoe & Vanilla PPO & DP-Shoe actor\\
 8 & PPO-DP-Adam & Vanilla PPO & DP-Adam actor\\
 \hline
\end{tabular}
\label{tab:policies}
\vspace{0.1cm}

\caption{List of policy classes used for experiments}
\end{table*}

\subsection{Reinforcement Learning}
We focus on non-deterministic environments with discrete and finite state and action spaces. Let $M = (S, A, P, R, \gamma, S_0)$ represent the MDP environment. Here, $S$ is the set of finite discrete states, $A$ is the set of finite actions, $P(s, a, s')$ is the transition probability of reaching state $s'$ by taking an action $a \in A$ in state $s$, where $s, s' \in S$. $R(s)$ is the reward the agent receives in state $s \in S$, $\gamma$ is the discount factor for future rewards, and $S_0$ is the initial state distribution over $S$. State value $V(s)$ is the value of expected return (sum of future discounted rewards) starting with state $s$. State-Action value $Q(s, a)$ denotes the value of taking action $a$ in state $s$ (following some policy $\pi$). The goal of an RL agent is to learn a policy that maximizes the expected cumulative reward. We use the following classes of RL algorithms to learn the optimal policy and value function for our experiments:
\begin{enumerate}
    \item VI \cite{pashenkova1996value}: VI computes an optimal state value function for an MDP. The method uses Bellman updates to converge to the optimal values.
    \item DQN \cite{mnih2013playing}: DQN is a model-free off-policy deep RL approach that uses a deep neural network for the Q-function which uses a batch of past experiences (replay memory) to train the agent to learn the optimal policy. 
    \item PPO \cite{schulman2017proximal}: PPO is a first-order optimization based policy-gradient algorithm that uses the actor-critic approach to find the best policy. The actor model learns to take an action in an observed state by improving upon the feedback given by the critic model - that takes state as an input, and finds a value function estimating future rewards with the help of a deep neural network. 
\end{enumerate}

\subsection{Inverse RL}
Inverse RL \cite{ng2000algorithms} is a method of extracting a reward function, given the observed, optimal behaviour in an environment. We use the method of inverse RL in finite-state spaces to reconstruct the private reward function by solving a linear programming (LP) \cite{ignizio1994linear} formulation that makes the given policy optimal by a large margin (as compared to other sub-optimal policies). Since this is an under-constrained problem, we choose the reward with the smallest $L_1$-norm.

\subsection{Differential Privacy}
DP \cite{10.1007/11681878_14} is considered to be the golden standard of computational privacy. It allows us to quantify the degree of privacy achievable by a mechanism. It is built on the concept of adjacent databases. In the context of our work, the RL agents learn optimal policies by exploring the environment and taking in rewards as a feedback for their actions. Since we care about the privacy of the rewards, we say that two reward functions are adjacent if the maximum $L2$ norm of their point-wise difference is upper bounded by $1$.
\begin{definition}
$(\epsilon, \delta)$-DP: A randomized mechanism $\mathcal{M}:\mathcal{D} \rightarrow \mathcal{R}$ with domain $\mathcal{D}$ and range $\mathcal{R}$ satisfies $(\epsilon, \delta)$-differential privacy if for any two adjacent inputs $d, d' \in \mathcal{D}$ and for any subset of outputs $\mathcal{S} \subseteq \mathcal{R}$ it holds that $$Pr[\mathcal{M}(d) \in \mathcal{S}] \leq e^{\epsilon} Pr[\mathcal{M}(d') \in \mathcal{S}] + \delta$$
\end{definition}

\begin{definition}
$\alpha$-R\`enyi Divergence: For two probability distributions P and Q defined over $\mathbf{R}$, the R\`enyi divergence of order $\alpha > 1$ is 
$$D_\alpha(P||Q) = \frac{1}{\alpha - 1} \log_e E_{x \sim Q} \left(\frac{P(x)}{Q(x)}\right)^\alpha$$
where P(x) and Q(x) are the respective probability densities of P and Q at x.
\end{definition}
An algorithm is said to have $(\alpha, \epsilon)$ R\`enyi DP \cite{mironov2017renyi} if for any two neighbouring databases, it holds that the R\`enyi divergence ($D_\alpha$) of order $\alpha$ between outputs of the algorithm is less than $e^\epsilon$.
\begin{definition}
$(\alpha, \epsilon)$-R\`enyi DP: A randomized mechanism $f:\mathcal{D} \rightarrow \mathcal{R}$ is said to have $(\epsilon)$-R\`enyi differential privacy of order $\alpha$, or $(\alpha, \epsilon)$-RDP for short, if for any adjacent $d, d' \in \mathcal{D}$ it holds that $$D_{\alpha}(f(d) || f(d')) \leq e^{\epsilon}$$
\end{definition}

\subsection{Private RL Methods}
We use the following private RL methods in our experiments: 
\begin{enumerate}
    \item \emph{Bellman update DP}: In this method, noise is added locally to the Bellman update step of VI, such that it satisfies the definition of $\epsilon$-DP \cite{10.1007/11681878_14}.
    \item \emph{R\'enyi-DP in DL}: This is a natural relaxation of DP that we use for multiple DP methods - DQN-DP-SGD, PPO-DP-SGD, and more \cite{abadi2016deep}, \cite{papernot2019making}.
    \item \emph{Functional noise DP in Q-Learning}: In this, functional noise is iteratively added to the value function in the training process. The aim is to protect the value function \cite{wang2019privacy}.
\end{enumerate}
We assume that we share the entire training process including every loss gradient update publicly, (worst-case privacy guarantee). 

\section{\pril: Privacy-Aware Inverse RL Analysis Framework}
\label{sec:4_inv_rl}

\begin{figure*}[h]
\centering
\includegraphics[scale=0.4]{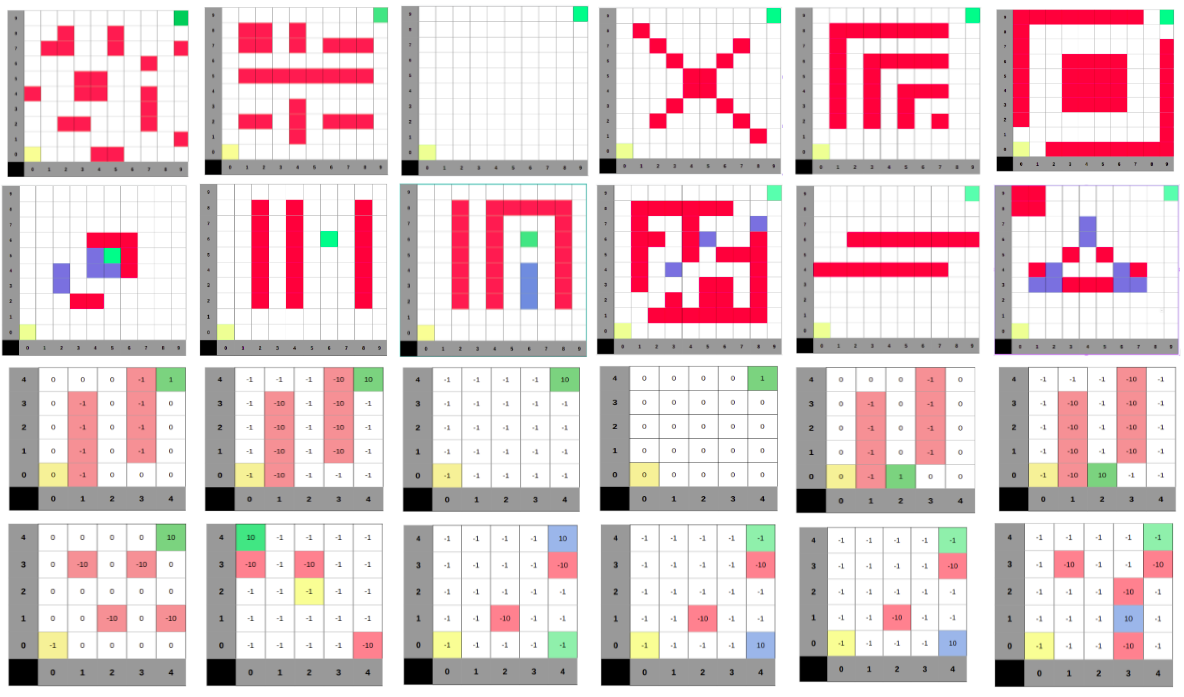}
\caption{All 24 FrozenLake environments used. Here, green: goal (G), red: frozen (F), yellow: start state, white: safe (S), and blue: high-reward (A).}
\label{fig:2_envs}
\end{figure*}
% TODO: Make this diagram consistent across grid sizes

We introduce a novel case of the data-reconstruction attack - the reward reconstruction attack for RL, as we wish to protect the reward function from adversaries. We assume that the adversary has knowledge of the environment and the learned private policy. Using this information, the adversary tries to reverse engineer the reward function. While many methods can be used to do so, we focus on the inverse RL technique in this paper, as it seems to be the best tool at our disposal. Using inverse RL, we perform the reward reconstruction attack, to determine how effective a private policy is at protecting the reward function. It does so by computing (a variety of) distances between the reconstructed reward and the original reward. The framework takes as input the original reward function $R$, an RL policy $P'$, and a private RL policy $P''$ trained using the same algorithm. Using the inverse RL algorithm, it predicts the reconstructed rewards, $R'$ and $R''$, from $P'$ and $P''$ respectively. It then computes the distances $d'(R', R)$ and $d''(R'', R)$, and compares them. The larger the distance, the stronger the RL policy's privacy guarantee (in protecting the reward function). We use multiple distance metrics, such as - \emph{$L_1$ norm, $L_2$ norm, $L_\infty$ norm, and number of sign changes}.
% TODO: Add citations for "While many methods can be used to do so"

\section{Experimental Setup}
\label{sec:5_exp_set_up}
\begin{figure*}[h]
\centering
\includegraphics[scale=0.32]{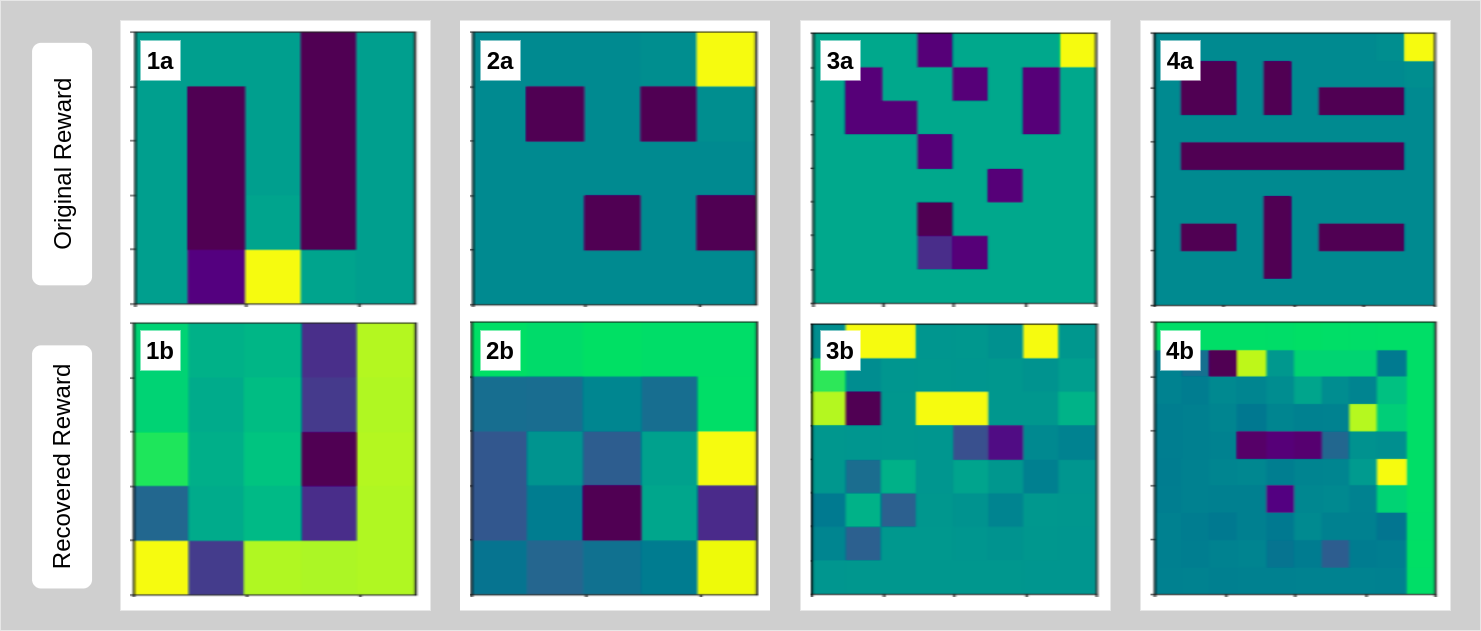}
\caption{Original and reconstructed reward heatmaps for two 5x5 FrozenLake environments (1a, 2a) and for two 10x10 FrozenLake environments (3a, 4a) }
% \kp{mention which algorithm} and what is "b" figure in each case
\label{fig:3_heatmaps}
\end{figure*}
% TODO: make the same figure as above for one grid for some technique with and without DP - qualitative comparison - 1a, 1b, 1c, 1d, ...

% \subsection{Overall Experimental Pipeline}
We will now discuss our overall experimental pipeline and setup. We perform our experiments on 24 custom environments (as shown in Figure \ref{fig:2_envs}) in the \emph{FrozenLake domain} - a discrete-state OpenAI Gym \cite{brockman2016openai} toolkit. In all these environments, the agent controls its movement and navigates in a grid-world. Additionally, the movement direction of the agent is uncertain and is only partially dependent on the direction chosen. The agent is rewarded for finding the most rewarding walkable path to the goal state. The grid-world environment has five possible states - safe (S), frozen (F), hole (H), high-reward (A) and goal (G). The agent has four possible actions - up, down, left and right. Half of the 24 environments are of a grid-size 5x5, and the remaining half are of a grid-size 10x10. The agent moves around the grid until it reaches the goal state. If it falls into the hole, it has to start from the beginning and is given a low reward. The process continues until it eventually reaches the goal state. We measure the performance of 8 private algorithms across three algorithm classes - \emph{VI, DQN, and PPO}:

For VI, we evaluate the performance of VI-DP-Bellman (private Bellman update via local DP) as well as non-private VI.

For DQN, we evaluate the performance of the following cases (with and without privacy): 
\begin{enumerate}
    \item \emph{DQN-DP-SGD}: DP-SGD optimizer + ReLU activations
    \item \emph{DQN-DP-Adam}: DP-Adam optimizer + ReLU activations
    \item \emph{DQN-DP-Shoe}: DP-SGD optimizer + tan-h activations 
    \item \emph{DQN-DP-FN}: DQN + functional noise
\end{enumerate}
For PPO, we evaluate the performance of the following cases (with and without privacy in the actor network): 
\begin{enumerate}
    \item \emph{PPO-DP-SGD}: DP-SGD optimizer + ReLU activations
    \item \emph{PPO-DP-Adam}: DP-Adam optimizer + ReLU activations
    \item \emph{PPO-DP-Shoe}: DP-SGD optimizer + tan-h activations
\end{enumerate}

\begin{figure}[!h]
\centering
\includegraphics[scale=0.35]{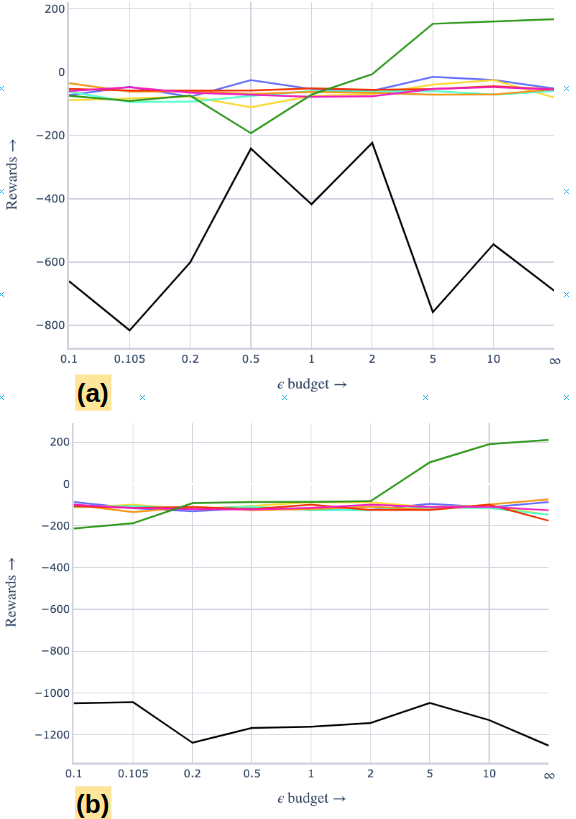}
\caption{Utility (test-time return) vs privacy trade-off for all policies averaged across grid-sizes 5x5 (a) and 10x10 (b). Legend is the same as that in figure \ref{fig:5_eps}.}
\label{fig:4_trade_off}
\end{figure}

We evaluate the \emph{privacy-utility trade-off} by simultaneously measuring the average returns of the the private policy over multiple sample trajectories during test time (as shown in Figure \ref{fig:4_trade_off}). For each private RL algorithm, we consider the non-private version of the RL policy (privacy budget $\epsilon = \infty$) as the baseline for reward reconstruction. The more private the policy is, the larger the reward distance should be (between the original reward and the reconstructed reward).

\emph{Reward distance as a measure of privacy guarantee}: We used four reward distance metrics for our experiments - \emph{$L_1$ distance, $L_2$ distance, $L_\infty$ distance, and the sign change count}. The idea is to measure the similarity between the original reward function and the recovered reward function. This is a measure of the degree of privacy of a policy - the larger the reward distance, the more private the policy is. The metrics are calculated as follows:
\begin{enumerate}
    \item $L_1$ distance: Normalize the rewards $R, R'$ using $L_1$-norm, and then take the $L_1$ distance across the 2 vectors.
    \item $L_2$ distance: Normalize the rewards $R, R'$ using $L_2$-norm, and then take the $L_2$ distance across the 2 vectors.
    \item $L_\infty$ distance: Normalize the rewards $R, R'$ using $L_\infty$-norm, and then take the $L_\infty$ distance across the 2 vectors. 
    \item Sign change count: Measure the number of sign changes from $R$ to $R'$.
\end{enumerate}
Since each distance metric is in a different space, all the distances evaluated together allow us to get a deeper insight into the reward reconstruction mechanism, and the optimality and privacy of policies.

\emph{Policy return as a measure of agent utility}: We measure how much utility the learned private policies achieve by observing how they perform during test-time, by calculating the average discounted returns over multiple trajectories played by the agent following the policy.

\begin{figure*}[!h]
\centering
\includegraphics[scale=0.27]{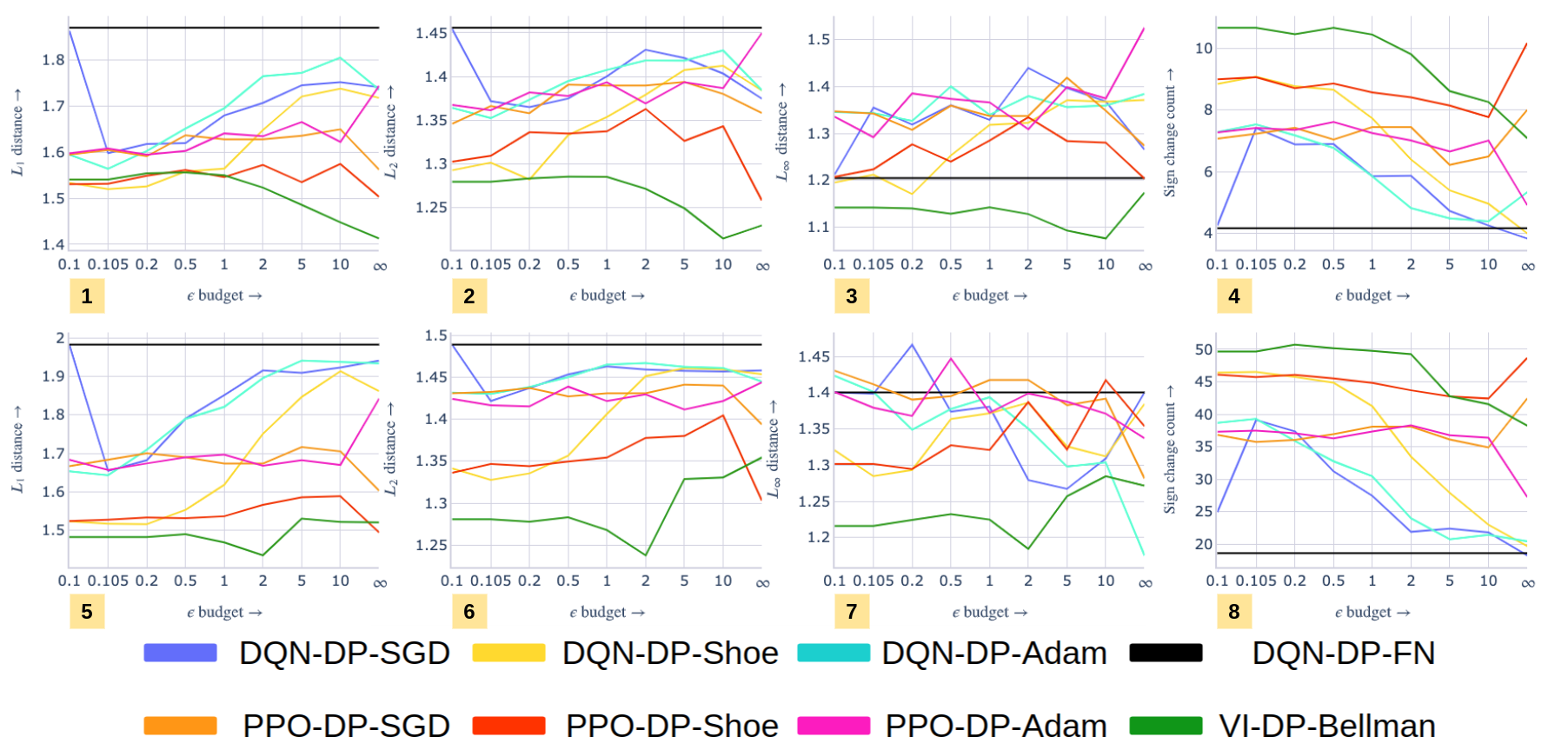}
\caption{Reward distance vs privacy budget graphs for all strategies: 1,5: $L_1$ distance, 2,6: $L_2$ distance, 3,7: $L_\infty$ distance, 4,8: Sign change counts. 1,2,3,4: averaged over 5x5 grid sized environments, 5,6,7,8: averaged over 10x10 grid sized environments}
\label{fig:5_eps}
\end{figure*}
% TODO: Show similar plots for the privacy (distance?) we are getting in Policy SPACE for different epsilon. This would bring to light that although there is expected privacy in Policy we have a random privacy for Reward reconstruction.

\emph{Implementation details}: We build 24 custom FrozenLake environments using the Open AI gym toolkit. We use an LP solver to solve the objective functions of Inverse RL. We build the Deep RL experiments using TensorFlow 2.4, and add privacy using TensorFlow Privacy. We build VI-DP-Bellman private algorithm from scratch, and use the publicly available code provided for the DQN-DP-FN strategy \cite{wang2019privacy}. We use a reference implementation \cite{alger16} for finite-state space inverse RL that makes the use of cvxopt \cite{andersen2013cvxopt} to solve the LP formulation. We use Linux OS based servers for training all the RL agents with a total of 8 GPUs and 8 CPUs. All experiments spanned across 9 privacy budgets, 24 environments, 8 policy classes, repeating each experiment 10 times (to account for the randomness stemming from private noise mechanisms and DL optimization). The total runtime for the entire set of experiments was 3 weeks. 

To the best of our knowledge, this is the first time such an attack and evaluation is conducted - hence the lack of pre-existing baselines. We evaluate each algorithm against the baseline case of no privacy ($\epsilon = \infty$). 
% We will share the entire documented code and data in the final version which will make our experiments easily reproducible. 

\begin{figure}[h]
\centering
\includegraphics[scale=0.4]{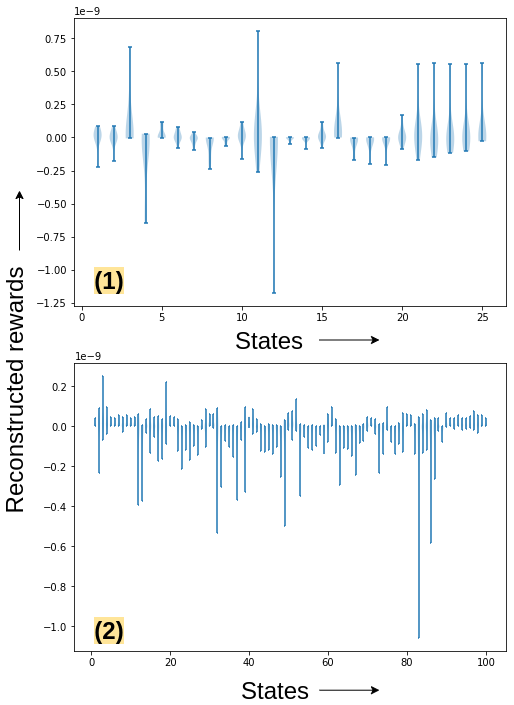}
\caption{Variance Violin plots for (1): PPO-DP-Shoe on an MDP of grid-size 5x5 with a privacy budget $\epsilon = 0.5$ showing variance over 10 runs; (2): VI-DP-Bellman on an MDP of grid-size 10x10 with a privacy budget $\epsilon = 0.5$ showing variance over 20 runs}
\label{fig:violin}
\end{figure}

\begin{table}[t]
\centering
\renewcommand{\arraystretch}{1.3} 
\begin{tabular}{|p{2cm}|p{3cm}|p{2cm}|}
%  \hline
%  \multicolumn{3}{|c|}{Sensitivity for Policy Classes} \\
 \hline
 No. & Policy Class & $l_2$-Sensitivity \\
 \hline
 1 & VI & 1.05 \\
 2 & DQN & 1.0 \\
 3 & PPO & 1.0 \\
 \hline
\end{tabular}
\label{tab:sens}
\vspace{0.3cm}
\caption{$l_2$ - Sensitivities of policy classes}
\end{table}

% We provide a detailed record of all DL and RL hyperparameters (such as learning rate, discount factor, convergence criteria, and more).
% We also list the $l_2$ sensitivities, and Gaussian noise standard deviations used corresponding to each privacy budget $\epsilon$, along with the formal privacy guarantees offered. 
\emph{Hyper-parameters and assumptions}:
\begin{itemize}
    \item We choose 9 different privacy budgets: - $\epsilon \in \{0.1, 0.105, 0.2, 0.5, 1.0, 2.0, 5.0, 10.0, \infty\}$ (including the no-privacy case where $\epsilon = \infty$). The way we choose these budgets is to have a diverse order of magnitudes within a reasonable range of budget values. We include $0.1$ and $0.105$ on purpose - to show the increased sensitivity of the private algorithm with slight changes to the budget at lower values. We observed a budget lower bound of $0.1$ - adding anymore noise results in a budget of $0$. We present the list of standard deviations used for each $\epsilon$ budget for each policy class in table 5.
    \item The intuition behind the DP-Shoe set of strategies comes from \cite{papernot2019making} titled "Making the shoe fit: : Architectures, initializations, and tuning for learning with privacy". Their key result states that SGD optimizer is strictly better than Adam optimizer, and that using tan-h activations are strictly better than using ReLu activations in the case of differentially private training. Hence, we named the DP-SGD optimizer with tan-h activations as the DP-Shoe method. This is also the reason why we do not perform experiments on the DP-Adam + tan-h combination.
    \item For all the DQN and PPO policy classes, we assume an $l_2$-sensitivity of $1.0$ - as all the private deep learning techniques allow for us to scale-down the data, to bring it to unit sensitivity. We present the list of $l_2$-sensitivities for each policy class in table 3. %\ref{tab:sens}.
    \item For PPO policy class, we assume that only the actor network will be using a DP optimizer (and not the critic network). We based this on the result from \cite{seo2020differentially}, which shows that DP-Actor outperforms DP-Critic and DP-Both (where both the networks are made private using DP).
    \item Learning rate $= 0.15$
    \item Mini-batch size = 50
    \item Number of micro-batches = 5 (each of size 10)
    \item Discount factor $\gamma = 0.99$
    \item Wind factor = 0.0001
    \item For reproducibility of the experiments, we fixed a random seed  and in our case we used seed = 0.
    \item Convergence threshold for value iteration $= 1.0 \times 10^{-10}$. Alternatively, we enforce the number of iterations to be less than 10000.
    \item For each graph in figure \ref{fig:5_eps}, the DQN-DP-FN strategy is giving straight lines. This implies that this strategy is extremely privacy agnostic in this domain. We made sure that there were no issues with the implementation itself. 
    \item Our method of creating grid-world environment maps aimed for diversity in spatial rewards and richness in reward structure. We built maps that would lead to interesting policies - while some policies would be very specific (like learning a single route), other policies would have multiple optimal routes.
    \item We also show that the absolute reward values themselves do not dictate the policy - it's their values taken relative to their surrounding reward values that influences the policy.
    \item We used the R\'enyi-DP privacy engine from the TensorFlow Privacy library to compute standard deviations (sigmas) for our chosen $\epsilon$ budgets, with an additive $\delta$ budget of $10^{-5}$ (standard default). 
\end{itemize}

\begin{table}[t]
\centering
\begin{tabular}{|p{0.5cm}|p{2cm}|p{1.25cm}|p{1.5cm}|p{1.5cm}|}
%  \hline
%  \multicolumn{5}{|c|}{DL Hyper-parameters for Policy Classes} \\
 \hline
 No. & Policy Class & Epochs & Iterations & Episodes \\
 \hline
 1 & VI & \ - & 10000 & 5\\
 2 & DQN & 15 & 200 & 5\\
 3 & PPO & 15 & 200 & 5\\
 \hline
\end{tabular}
\label{tab:hyper}
\vspace{0.3cm}
\caption{DL hyper-parameters for testing}
\end{table}

\section{Analysis and Discussion}
\label{sec:6_discuss}
Figure \ref{fig:5_eps} presents the variation in reward distances (y-axis) (of each type: $L_1, L_2, L_\infty$, and sign change counts; from the original reward) with an increase in the $\epsilon$ privacy budget (x-axis) (9 discrete values) including the no-privacy case at the very end ($\epsilon = \infty$). The first row shows results averaged over the 12 FrozenLake environments of grid size 5x5, whereas the second row shows the results corresponding to the environments of grid size 10x10. Each graph shows this relationship for all 8 private algorithms: DQN-DP-SGD, DQN-DP-Shoe, DQN-DP-Adam, DQN-DP-FN, PPO-DP-SGD, PPO-DP-Shoe, PPO-DP-Adam, and VI-DP-Bellman. The graphs show that there is no clear indication of any private strategy improving at reconstructing the reward (wrt all distances) with a relaxation in the privacy budget - thus, rendering all strategies ineffective at being a truly meaningful private strategy. The reward reconstruction distance(s) are independent from the privacy budgets across all policy classes. We observe the same lack of trend across both rows, i.e., both for the 5x5 results in row 1 and 10x10 results in row 2. 

Figure \ref{fig:4_trade_off} presents the trade-off between the amount of utility (expected return: y-axis) and the degree of privacy ($\epsilon$ budget: x-axis) achieved by a private RL privacy. Graph 1 gives us the average trade-off for 5x5 environments, and graph 2 - for the 10x10 environments. Almost all algorithms exhibit comparable performance - with the exception of DQN-DP-FN, which performs significantly worse. 

\begin{figure}[h]
\centering
\includegraphics[scale=0.3]{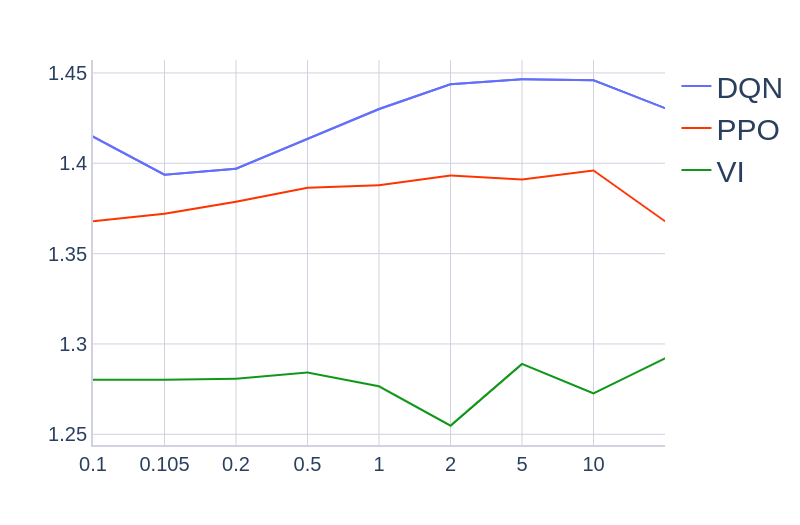}
\caption{Aggregated $L_2$ distances across all environments and policy variants of the three main classes of RL algorithms: DQN, PPO, and VI.}
\label{fig:6_aggregate}
\end{figure}

Figure \ref{fig:3_heatmaps} shows the heatmaps and reward structures of 4 different MDPs (row 1) and their corresponding reconstructed rewards (row 2) using the VI-DP-Bellman algorithm. From 1a and 1b we can observe that the agent is able to clearly detect the obstacle bar on the right. From 2a and 2b we can infer that the agent finds a straight line path along the edges that is rewarding. But it also wrongly identifies some states as rewarding, even though they might be dead-ends (bright yellow bottom-right corner). Perhaps, this could be because the agent is able to survive the cost of the state directly above it, if it means that the agent can easily reach the goal state in a short amount of time. In 3a and 3b we learn that even in such a rich environment, the agent is able to reconstruct the reward structure from an implementation of the policy. It can clearly identify rewarding blocks within the entire maze. And finally, in 4a and 4b, the agent learns to stick to the edges to maximize reward early-on - which is why its evaluation of the central region is poor, and privacy preserving.

Figure \ref{fig:6_aggregate} shows that DQN algorithms give us the strongest reward privacy, followed by PPO algorithms, followed by VI algorithms - that do a very poor job at protecting the reward - despite having very similar utilities (from figure \ref{fig:4_trade_off}).

Figure \ref{fig:violin} presents violin plots for specific instances of PPO-DP-Shoe and VI-DP-Bellman techniques on an MDP of size 5x5 and 10x10 respectively, averaged over multiple runs. The x-axis shows all the states whereas the y-axis shows violin plots of the value of the reconstructed reward. It shows that across multiple runs of our experiments, we are able to bound the reconstructed reward's variance (for each state) by a term of the order of $10^{-19}$. We observe similar results in most other cases as well. These rewards are learnt by an adversary via the \pril \ framework. 

Based on the experiments performed, we can say that there is a considerable gap between the privacy provided by the existing private methods (in policy), and the level of privacy needed to protect the reward function from the inverse RL attack. We can also infer that techniques using Deep RL methods are able to learn the policy in a more general manner, as compared to non-deep methods. We address the need for better privacy techniques for RL algorithms that can effectively protect the reward function. We hope that our work inspires a deeper theoretical understanding of the limits to minimizing the gap, as well as its consequences in real-world applications. Besides thoroughly testing our code, we perform an extensive span of experiments. Contrasting our results with the baseline (no privacy), we find the reward distances to be quite similar. We therefore, believe that the source of the privacy gap is not experimental error. Our survey of papers that experiment on FrozenLake shows that the commonly used grid sizes are $\{4\times4, 8\times8\}$ while we experimented with slightly larger grid sizes - $\{5\times5, 10\times10\}$. We expect to observe a similar (or worse) privacy gap upon further increase of grid size since the reward would be richer in information, and the DP sensitivity is independent of the grid size. 

% PRIL only giving comparative description and having a threshold for budget and reward distance: Selecting a budget threshold and reward distance threshold is subjective - it depends on how sensitive the information is and the agent's mode of operation. The budget threshold can act as an elbow point - at what budget do we obtain a good trade-off between privacy and reward distance? The reward distance can be calibrated against the baseline case ($\epsilon = \infty$) e.g., to ensure that the distance does not become any smaller than $x(=10) \%$ of the baseline distance. We appreciate the concerns raised and will address it in the final version.
% Privacy Budget Analysis: Improvement in privacy-utility trade-off for more generalized PPO policy using Network Randomization % Policy Stability: Since query sensitivity is the same as stability of a DL model, making the model private makes it more stable % Limitations

While we demonstrate our work on a grid-world domain, we believe it is extendible to real-world domains with sensitive data. Our work is the first in this direction and serves as evidence that there is a need to inspect further. Deep RL is increasingly being used for recommendation systems (RecSys) in dynamic environments. Consider the case when the recommendation engine for every user is a unique private RL policy whose job is to recommend items to users and learn their preferences in an online fashion (given the user’s historical data). The reward is the user’s feedback (ratings) to the recommended action. While the policy provides privacy guarantees for its training process, it can leak the user’s feedback when subject to the re-identification attack via reward reconstruction. \pril can help assess this threat better.

We surveyed a range of Inverse RL (IRL) algorithms - finite state space LP, sample trajectories (\cite{ng2000algorithms}), deep IRL (\cite{wulfmeier2015deep}), and maximum entropy IRL (\cite{ziebart2008maximum}, \cite{wulfmeier2015maximum}). Despite starting with the simplest case - LP for finite state spaces, we observe a significant privacy gap. The LP method acts as a baseline for other IRL methods. With increased complexity, the reward function would be represented parametrically which would allow the system to evaluate performance on much larger and richer (and potentially continuous) environments. As the performance of IRL as an attacker improves, we expect the issue of privacy gap to become even more important to address.

Our work introduces a novel direction of evaluating the privacy guarantees of RL systems. In the future, we hope to build on our work in multiple ways: extending to the multi-agent scenario, extending to a diverse set of domains, assessing the effect of generalization and exploration on privacy, testing the performance of other RL algorithms such as PPO-Clip and PPO-KL, and evaluating the performance of other complex inverse RL algorithms in improving the framework.

\section{Conclusion}
\label{sec:7_conclusion}
This paper introduces a new Privacy-Aware Inverse RL analysis framework (\pril) for enhancing reward privacy in reinforcement learning (RL) that performs a novel reward reconstruction attack and demonstrated its ability to fairly assess the level of privacy achieved in protecting the reward structure from adversarial attacks. We studied the set of existing privacy techniques for RL, performed a detailed evaluation of their effectiveness and identified that there is a significant gap between the current standard of privacy offered and the standard of privacy needed to protect reward functions in RL. We quantify this gap by measuring distances between the original and reconstructed rewards via the reward reconstruction attack.
% TODO: Add possibility of future work? Urge community to focus on reward privacy as well ****** 

\begin{table*}[t]
\centering
\begin{tabular}{|p{1cm}|p{1.25cm}|p{1.5cm}|p{1.25cm}|p{1.25cm}|p{1.25cm}|p{1.25cm}|p{1.25cm}|p{1.25cm}|p{1cm}|}
%  \hline
%  \multicolumn{11}{|c|}{Sigmas for Policy Classes} \\
 \hline
 Policy & $\epsilon =$ 0.1 & $\epsilon =$ 0.105 & $\epsilon =$ 0.2 & $\epsilon =$ 0.5 & $\epsilon =$ 1.0 & $\epsilon =$ 2.0 & $\epsilon =$ 5.0 & $\epsilon =$ 10.0 & $\epsilon = \infty$ \\
 \hline
 VI  & 2080.08   & 1886.69 & 520.02  & 83.20  & 20.80 & 5.20  & 0.83 & 0.21 & 0        \\
 DQN & 94229     & 150     & 22.75   & 9.89   & 5.38  & 3.03  & 1.55 & 1.0  & 0        \\
 PPO & 94229     & 150     & 22.75   & 9.89   & 5.38  & 3.03  & 1.55 & 1.0  & 0        \\
 \hline
\end{tabular}
\label{tab:sigmas}
% \vspace{0.01cm}
\caption{Standard deviations}
\end{table*}

\bibliography{refs.bib}
\bibliographystyle{plain}
\appendix
\section{Appendix}

Here we will present the scope of our work along with the sensitivity proof of our VI-DP-Bellman strategy. 

\subsection{Scope}
While there are multiple privacy techniques for RL algorithms, many of them are not directly relevant for the work done in this paper.

\begin{itemize}
    \item We decided not to adopt the private upper confidence bound (UCB) and private deep Q learning methods from \cite{vietri2020private}, as their mode of operation is completely different. They assume that every episode of experience comes from a different private entity, whereas we focus on protecting any single reward function.
    \item We decided not to apply the DP-FN technique to PPO as it is not a natural extension. It would require significant deeper analysis to be able to apply DQN's private weight update to the updates to the actor network in PPO, considering that the interleaved updates to the critic network would be a source of privacy leakage. 
    \item In the space of inverse RL, we used the best solution for our domain - where we have access to the transition dynamics for a finite state space. For future work, we will look at how other inverse RL techniques fare in large state space domains, or in domains where we do not have access to the transition dynamics. 
\end{itemize}

\subsection{VI-DP-Bellman Strategy}
Value Iteration with output DP (global DP) allows us to publish only the final state values. However, if we want the entire learning process and each update to be public, we need to add noise to the Bellman update step. The key step is to add Gaussian Noise to the Bellman update step as follows: 

\begin{equation}
    V(s) = \max_a \Sigma_{s'} T(s,a,s') [r(s,a) + \gamma V(s')] + \mathcal{N}(0, \sigma)
\end{equation}

Here, the $\sigma$ parameter of the Gaussian noise distribution (standard deviation) is computed from the privacy budget $\epsilon$ and the sensitivity of $f$: $\delta f$, using the R\'enyi DP guarantee for Gaussian noise mechanisms. We now calculate the sensitivity of the Bellman update.

\paragraph{Sensitivity Proof}
The sensitivity of the state-value update via the Bellman equation is upper bounded by $(n+1)/n$, where $n = |S|$ is the number of states in the discrete environment.
We consider two reward functions $r, r'$ to be neighbouring reward functions if $||r - r'||_2 \leq 1$. For the $r(s)$ reward structure, the gain can be represented as:

The gain at time-step $t$ from all future rewards until time-step $t+T+1$ is:

\begin{equation}
   G_t  = \Sigma_{j=0}^T \gamma^j r_{t+j+1} 
\end{equation}
The state-value function $V_\pi(s)$ of an MDP is the expected return starting from state $s$, and then following policy $\pi$.
\begin{equation}
V_\pi(s) = E_\pi[G_t | s = s_t] 
= E_\pi [\Sigma_{j=0}^T \gamma^j r_{t+j+1} | s = s_t]
\end{equation}
Bellman equation for optimal value function:
\begin{equation}
    V(s) = \max_a \Sigma_{s'} P(s,a,s') [r(s,a) + \gamma V(s')]
\end{equation}

Let us consider two Value functions $V_1, V_2$, each corresponding to a different reward function $r_1, r_2$ for the same environment such that $||r_1 - r_2||_2 \leq 1$.

\begin{equation}
    V_1(s) = \max_a \Sigma_{s'} P(s,a,s') [r_1(s,a) + \gamma V(s')] 
\end{equation}
\begin{equation}
    V_2(s) = \max_a \Sigma_{s'} P(s,a,s') [r_2(s,a) + \gamma V(s')] 
\end{equation}

Let the difference value function be $\nabla V = V_1 - V_2$ and reward function be $\nabla r = r_1 - r_2$.

\begin{equation}
    \begin{aligned}
      \nabla V (s) &= \max_{a_1} \Sigma_{s'} P(s,a_1,s') [r_1 (s,a_1) + \gamma V_1 (s')] \\
              &  - \max_{a_2} \Sigma_{s''} P(s, a_2, s'') [r_2(s,a_2) + \gamma V_2(s'')] \\
    \end{aligned}
\end{equation}

We can assume that $V_1$ is estimated by following policy $\pi_1$ learnt from $r_1$, and similarly, $V_2$ is estimated by following policy $\pi_2$ learnt from $r_2$. Then, 

\begin{equation}
\begin{aligned}
  \nabla V (s) &= \Sigma_{a_1} \pi_1(a_1|s) \Sigma_{s'} P(s,a_1,s') [r_1 (s,a_1) + \gamma V_1 (s')] \\
 & - \Sigma_{a_2} \pi_2(a_2|s) \Sigma_{s''} P(s, a_2, s'') [r_2(s,a_2) + \gamma V_2(s'')] \\
\end{aligned}
\end{equation}

Upon re-arranging the terms, 

\begin{equation}
    \begin{aligned}
      \nabla V(s) &= \Sigma_{a} \Sigma_{s'} P(s,a,s') [\pi_1(a|s) ( r_1(s,a) + \gamma V_1(s')) \\
      & - \pi_2(a|s) (r_2(s,a) + \gamma V_2(s'))] \\
    \end{aligned}
\end{equation}

\begin{equation}
    \begin{aligned}
        \nabla V(s) &= \Sigma_{a} \Sigma_{s'} P(s,a,s') [(\pi_1(a|s) r_1(s,a) - \pi_2(a|s) r_2(s,a)) \\
        & + (\gamma \pi_1(a|s) V_1(s') - \gamma \pi_2(a|s) V_2(s'))] \\
    \end{aligned}
\end{equation}

To simplify,

\begin{equation}
\begin{aligned}
        \nabla V(s) &= \Sigma_{a} \Sigma_{s'} P(s,a,s') [(\pi_1 r_1 - \pi_2 r_2) \\ 
        & + \gamma (\pi_1 V_1 - \pi_2 V_2)] \\
\end{aligned}
\end{equation}

We know that $0 \leq P(s,a,s'), \pi(a|s), \gamma \leq 1$. Therefore, we can always say that $\Sigma_{a} \pi(a|s) r(s,a) \leq \max_{a} r(s,a)$. In our case, we use $r(s)$, so $\max_{a} r(s,a) = r(s)$. Similarly, we can use the same logic to resolve $P(s,a,s')$ and $\gamma$ in the expression. So, we can upper bound $\nabla V(s)$ as follows: 

\begin{equation}
    \nabla V(s) \leq (r_1 - r_2)(s) + \Sigma_{s'}(\max V_1 - \min V_2)
\end{equation}

%%%%%%%%%%%%%%%%%%%%%%%%%%%%%
\begin{equation}
    \nabla V(s) \leq (r_1 - r_2)(s) + \Sigma_{s'} \max (V_1 - V_2)
\end{equation}

\begin{equation}
    \nabla V(s) \leq (r_1 - r_2)(s) + \Sigma_{s'} \max \nabla V(s')
\end{equation}

\begin{equation}
    \nabla V(s) \leq (r_1 - r_2)(s) + \Sigma_{s'} || \nabla V(s') ||_\infty
\end{equation}

Since $l_\infty$-norm is always $\leq l_2$-norm, 

\begin{equation}
\begin{aligned}
    \nabla V(s) & \leq (r_1 - r_2) + \Sigma_{s'} || \nabla V(s') ||_\infty \\
     & \leq (r_1 - r_2) + \Sigma_{s'} || \nabla V(s') ||_2 \\
\end{aligned}
\end{equation}

Taking $l_2$-norm of the entire equation,
\begin{equation}
\begin{aligned}
    ||\nabla V(s)||_2 & \leq || ((r_1 - r_2) \\
        & + \Sigma_{s'} || \nabla V(s') ||_2) ||_2 \\
\end{aligned}
\end{equation}

Let $(r_1 - r_2) (s) = \nabla r(s)$, and $\Sigma_{s'} \nabla V(s') = \nabla W$. Notice that $\nabla W$ is independent of (any) state $s$. Then,

\begin{equation}
\begin{aligned}
     & || ((r_1 - r_2)  + \Sigma_{s'} || \nabla V(s') ||_2) ||_2 \\
     & = ((\nabla r(s) + \nabla W)^2)^{0.5} \\
     & = \nabla r(s) + \nabla W \\
\end{aligned}
\end{equation}

Thus, 
\begin{equation}
    \nabla V(s) \leq (r_1 - r_2)(s) + \Sigma_{s'} \nabla V(s')
\end{equation}

If we sum the above term across all states,

\begin{equation}
\begin{aligned}
     \Sigma_{s} \nabla V(s) & \leq \Sigma_{s} (r_1 - r_2)(s)\\
    & + |S| \times (\Sigma_{s'} \nabla V(s'))\\
\end{aligned}
\end{equation}

\begin{equation}
\begin{aligned}
    \Sigma_{s} \nabla V(s) & \leq \Sigma_{s} (r_1 - r_2)(s) \\
     & + |S| \times (\Sigma_{s'} \nabla V(s')) \\
\end{aligned}
\end{equation}

Now the LHS is the same as the second term in the RHS. So, by shifting terms around, 
\begin{equation}
\begin{aligned}
     \Sigma_{s} \nabla V(s') & \leq \frac{\Sigma_{s \in S} (\nabla r)(s)}{|S-1|}\\
\end{aligned}
\end{equation}

Considering $||\nabla r||_2 \leq 1$, we can generalize that 
$$||\Sigma_{s \in S} (\nabla r)(s)||_2 \leq |S| . 1$$

Therefore, 
\begin{equation}
     \Sigma_{s} \nabla V(s') \leq \frac{|S|}{|S-1|}
\end{equation}

Or, $$||V_1 - V_2||_2 \leq \frac{|S|}{|S-1|}$$.

Therefore, we are able to show that the $l_2$-sensitivity of the Bellman Value update for Value Iteration is $\nabla V \leq \frac{|S|}{|S-1|}$, where $|S|$ is the number of states. For our domains, the largest sensitivity was $\frac{25}{24} = 1.041$, which we upper bounded with a value of 1.05.

\end{document}